\documentclass[runningheads]{llncs}
\usepackage{amsmath,graphicx}
\usepackage{amsfonts,amssymb,amsmath}
\usepackage{fancyhdr}
\newcommand*{\affmark}[1][*]{\textsuperscript{#1}}
\usepackage{graphicx}
\begin{document}
\title{re-OBJ: Jointly learning the foreground and background for object instance re-identification}
\authorrunning{Vaibhav Bansal, Stuart James and Alessio {Del Bue}}
\titlerunning{re-OBJ}
%
\author{Vaibhav Bansal\affmark[1,3], Stuart James\affmark[2] and Alessio {Del Bue}\affmark[1]}
%
%
\institute{Visual Geometry and Modelling (VGM), Istituto Italiano di Tecnologia \affmark[1] \\ Center for Cultural Heritage Technology (CCHT), Istituto Italiano di Tecnologia\affmark[2] \\  
Universit\`{a} degli studi di Genova\affmark[3]}

\maketitle              
\begin{abstract}
Conventional approaches to object instance re-identification rely on matching appearances of the target objects among a set of frames. However, learning appearances of the objects alone might fail when there are multiple objects with similar appearance or multiple instances of same object class present in the scene.
This paper proposes that partial observations of the background can be utilized to aid in the object re-identification task for a rigid scene, especially a rigid environment with a lot of reoccurring identical models of objects. 
Using an extension to the Mask R-CNN architecture, we learn to encode the important and distinct information in the background jointly with the foreground relevant to rigid real-world scenarios such as an indoor environment where objects are static and the camera moves around the scene.
We demonstrate the effectiveness of our joint visual feature in the re-identification of objects in the ScanNet dataset and show a relative improvement of around $28.25\%$ in the rank-1 accuracy over the deepSort method.

\end{abstract}
\section{Introduction}
\begin{figure*}
    \centering
    \includegraphics[width=0.8\textwidth]{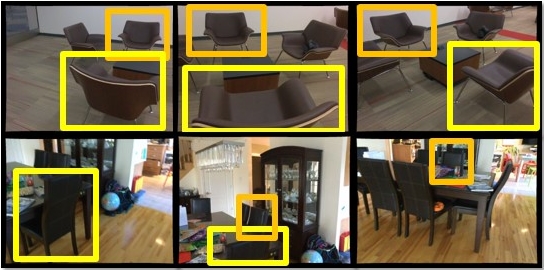}
    \caption[width=0.5\textwidth]{Similar looking objects in rigid, indoor scenes from ScanNet dataset. Multiple instances of the same object class, chair, in this case, are hard to differentiate with each other. In such cases, background can be highly useful to re-identify a particular instance in multiple views.}
    \label{fig:examp1}
\end{figure*}

Multiple object matching and association are classical problems in many important tasks such as video surveillance, semantic scene understanding and also, Simultaneous Localization And Mapping (SLAM). Given an indoor scene, where the environment is frequently cluttered with several near-identical objects, it is challenging to identify and track a particular instance of an object among a number of objects present in the scene, e.g. see Fig.~\ref{fig:examp1}. 
The problem is even more challenging when there is a wide baseline among multiple 
views (or temporally disjoint). It is complex to re-identify a vast variety of objects based on appearance only. There are many challenges for the association problem i.e. 
occlusions, motion blur, mis-detections, etc. Conventional methods use two major approaches to build a re-ID system - appearance-based and motion-based. Most methods use an appearance-based approach because motion prediction based systems try to localize each object instance based on a motion model, however, due to the possibility of huge unpredictable trajectories across the frames, these methods tend to fail when the same object instance reappear after a long time. 

Many previous studies focus on \textit{person} re-identification where the goal is to assign a correct ID of an instance of a specific class (i.e. a pedestrian) across multiple-views obtained from cameras with possibly non-overlapping views. In general, these methods try to learn discriminative features based on person's face~\cite{schroff2015facenet}, clothing~\cite{li2015clothing} or symmetry-driven local features~\cite{farenzena2010person} to re-ID people. In contrast, the problem of associating an unique ID to instances of objects is often solved as the association of multiple unknown objects between views \cite{milan2016mot16}. This problem is closely related to person re-ID and often evaluated in the pedestrian (person) scenario with early work on PET2009 \cite{conte2010performance}.  


However, the specific task of re-identifying multiple near-identical objects in a rigid scene presents a different challenge, we refer to as re-OBJ, a specific case of re-ID.
In this paper, we consider a static indoor video dataset where large displacement in the camera motion is unlikely and so the background of an instance cannot undergo a sudden drastic change. Therefore, we propose to jointly learn the foreground and the background to build a robust object re-identification system at the instance level. We propose not only to learn the appearance of an object but also the background that can provide a lot of useful information regarding the surroundings of an instance which is unique to that instance at any given viewpoint. Consider a scene of an office room with multiple chairs and tables present. To re-identify a particular object instance across multiple images, it is important to be able to distinguish it from other instances of the same object class. Intuitively, if we can observe and encode the surroundings of that particular instance within a stream of images, we can be confident to an extent that the object instance in consideration has been seen before and it is different from other instances of the same class because the environment around it is unique at any given point of time even when other instances have similar appearance. (see Figure~\ref{fig:examp1}). 


\section{Related work}

There is a vast literature for object re-identification that is mostly focused on person re-identification. The ability to re-identify objects in the images heavily relies on finding a similar set of images for a given image of the target object, possibly with multiple instances, using visual search to retrieve similar images to the given query image. Some works in the literature like~\cite{bhuiyan2015exploiting,farenzena2010person} exploit the knowledge that the same individual is been detected in consecutive frames and then learning an appearance-based transfer function for a robust re-identification system. Additionally, in~\cite{farenzena2010person}, they extract features from three different complementary modalities: the chromatic content, spatial arrangement of colors and local motifs derived from different parts of the human body to accumulate local features. 
Other deep learning models learn the category-level similarity~\cite{taylor2011learning} that mainly involves semantic similarity. 
The study highlights the effect of significant visual variability within a category although the semantic and visual similarities are generally quite consistent across different categories. Thus, applications that involve the computation of image similarity like re-identification, image retrieval, search-by-example require learning a fine-grained image similarity that can also distinguish the differences between different images of the same category. Relative attribute~\cite{parikh2011relative} learns image attribute ranking among the images with the same attributes. OASIS~\cite{chechik2010large} performs local distance learning~\cite{frome2007image} learn image similarity ranking models on top of the hand-crafted features. Such appearance-based approaches are good at distinguishing intra-class variation, in contrast, we focus on the objects' relationship to the background to jointly learn a foreground and background discriminative appearance feature.

Many image similarity models~\cite{boureau2010learning,chechik2010large,taylor2011learning} simply extract features like Gabor filters, SIFT~\cite{lowe1999object}, HOG~\cite{dalal2005histograms} features to learn similarity between images. However, the representation of the hand-crafted features limits the performance of these methods. Some deep learning-based models popular in image classification tasks~\cite{krizhevsky2012imagenet} have shown great success in learning features from the images but these models cannot directly fit similar image ranking especially the fine-grained distinction between similar images. Thus, in order to learn the fine-grained image similarity deep ranking model has been proposed by~\cite{wang2014learning}. Pairwise ranking model is a widely used learning-to-rank formulation. It is used to learn image ranking models in~\cite{chechik2010large,frome2007image,parikh2011relative}. Generating good triplet samples is a crucial aspect of learning pairwise ranking model. FaceNet~\cite{schroff2015facenet} showed that the triplet loss is a suitable loss function for the verification, recognition and clustering than the verification loss~\cite{schultz2004learning}. The difference is that the verification loss minimizes the $L2$-distance between objects of the same identity and enforces a margin between the distance of objects of different identities whereas the triplet loss also encourages a
relative distance constraint and thus, enhancing the ability to discriminate between dissimilar identities. A closely related work, RIO~\cite{Wald2019RIO}, introduced a method for object instance re-localization in 3D. They use a fully-convolutional 3D correspondence network to find matching features related to multiple objects in changing 3D scans in order to estimate their corresponding 6DoF poses in another scan of the same indoor environment differed by time. In~\cite{chechik2010large} and~\cite{parikh2011relative}, the triplet sampling algorithms assume that we can load the whole dataset into memory, which is impractical for a large dataset. Our work is built upon the deep ranking model proposed by~\cite{wang2014learning} with an efficient triplet sampling algorithm that does not require loading the whole dataset into the memory.

\section{Object instance separation encoding}

\begin{figure*}[ht]
    \centering
    \includegraphics[width=0.7\textwidth]{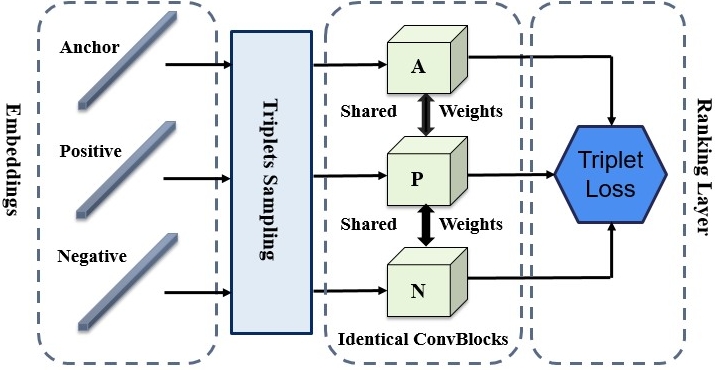}
    \caption[width=0.5\textwidth]{Triplet of input tensors corresponding to images. Each tensor contains an embeddings of the anchor image A, positive image P and a negative Image N which are fed into three identical deep neural networks independently with shared weights where the triplet loss is optimized.}
    \label{fig:arch2}
\end{figure*}

For a robust object re-identification system for a rigid scenario, we hypothesize that the background information is useful in order to discriminate between multiple instances of the same semantic class and also the objects that have a similar appearance as shown intuitively in Figure~\ref{fig:examp1}. To include the background information, the first step in our approach is to use an off-the-shelf object detector, i.e. Mask-RCNN (sec.~\ref{sec:method_detection}), and obtain foreground masks of the objects with the bounding boxes that are expanded (see sec.~\ref{sec:training_data}) in order to include a substantial background around the object within the bounding boxes. Encodings from the separated masked foregrounds and the masked backgrounds are extracted using ResNet50 (sec.~\ref{sec:method_encoding}), which are concatenated to obtain joint embeddings. These embeddings then are sampled into triplets $\{positive, negative, anchor\}$ and fed to a triplet-based network architecture consisting of three identical ConvNets (see Figure~\ref{fig:arch2} and Figure~\ref{fig:arch3}) with the pairwise ranking model to learn image similarity for a triple-based ranking loss function.    

\subsection{Object Detection}
\label{sec:method_detection}
\begin{figure*}[!ht]
    \centering
    \includegraphics[width=0.85\linewidth]{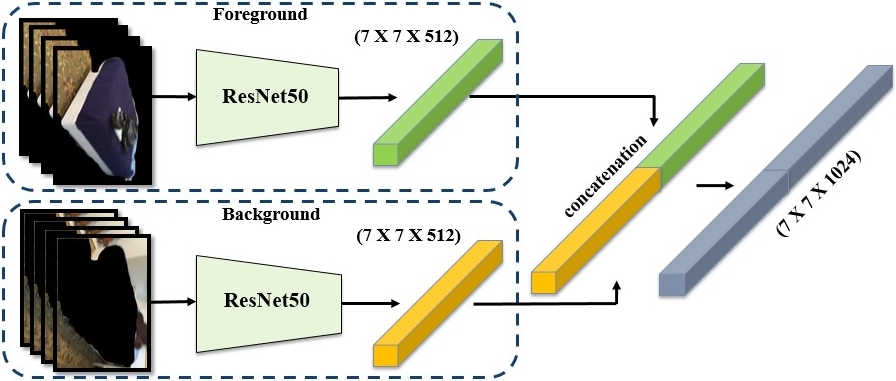}
    \caption[width=0.2\linewidth]{As input, our network takes expanded bounding boxes (see sec.~\ref{sec:method_encoding}) which construct a pair of images for masked foreground and masked background (seen on left of image). Each of the pair of images is passed through a ResNet50 where we take an intermediary representation $7\times7\times512 $ providing spatial information, which is concatenated to provide a joint representation of $7\times7\times1024$. 
    
    }
    \label{fig:my_arch}
\end{figure*}

\begin{figure*}
    \centering
    \includegraphics[width=0.9\textwidth]{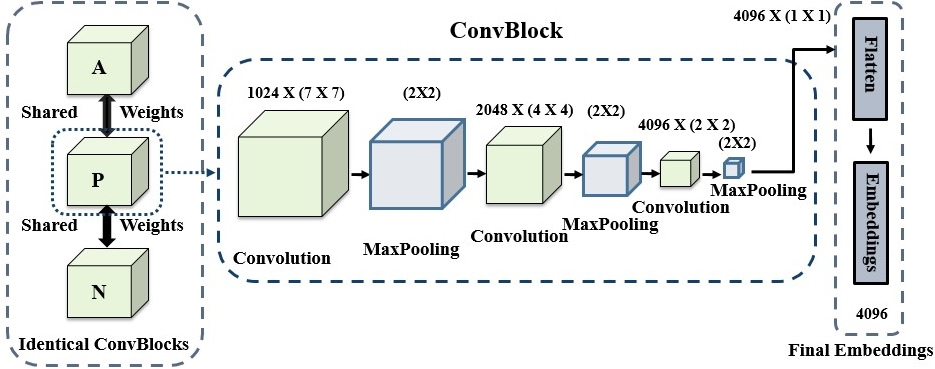}
    \caption[width=0.5\textwidth]{Our ConvBlock takes in the encoding from fig.~\ref{fig:my_arch}. The ConvBlock consists of a network of convolutional and maxpooling layers, which pool the spatial information and merge the foreground and background encodings to obtain final embeddings.}
    \label{fig:arch3}
\end{figure*}

Our approach relies on previous work, Mask-RCNN~\cite{he2017mask} which uses region-based object detector like Faster R-CNN to detect objects. It does not only provide a bounding box around an object but also performs image segmentation and provides a mask representing a set of pixels belonging to the same object. A Region Proposal Network (RPN) is used to generate a number of region proposals followed by a position-sensitive RoI pooling layer to warp them into a fixed dimension. Finally, it is fed into fully-connected layers to produce class scores and the bounding box predictions. A parallel branch of two additional fully-connected layers provides the mask. 
Using the output from the Mask-RCNN, we extract each bounding box including masks as separate images and resize them into images of a fixed size in order to train our network to learn a visual encoding of the objects' mask and the background surrounding them within the bounding boxes (see first column, Figure~\ref{fig:my_arch}).




\subsection{Object Visual Encoding}
\label{sec:method_encoding}
For each object of the input images, we create two sets of images $F=\{I_{f}, I_{b}\}$. Using the detections obtained from Mask-RCNN, one set is created by extracting masks representing objects in the foreground ($I_{f}$). The other set only contains the background with the subtracted foreground ($I_{b}$). As shown in Figure~\ref{fig:my_arch}, a pair of images is taken from each set to pass through two identical streams to learn an encoding between the masked foreground and the background. 
Each of the images, the masked background and the masked foreground is input to a ResNet50~\cite{he2016deep} deep model pre-trained on ImageNet~\cite{deng2009imagenet} dataset to extract the features. We take from an intermediary layer of the network providing $I_{(.)} \in \mathbb{R}^{7\times7\times512}$ representation of the two images retaining spatial context, the tensors are then concatenated to provide an embedding $F \in \mathbb{R}^{7\times7\times1024}$.


\subsection{Triplet loss}
\label{sec:method_association}
An effective algorithm for object instance re-identification should be able to distinguish not only between the images of different objects but also between different instances of the same object class. Especially, in the indoor scenes where multiple instances of the same object category are present, i.e.\ an office with multiple tables and chairs; it is highly challenging to re-identify a particular object instance amongst others.

A triplet of images has three kinds of images: an \textit{anchor} which acts like a query template, a \textit{positive} and a \textit{negative} image. In order to ensure an effective re-identification at the instance level, it is important to also consider the intra-class variations and different instances of the same object as negative examples. For example, a backpack and a chair are definitely an example of \textit{anchor-negative} pairs but two different instances of the same chair (with a different background) should also be considered an~\textit{anchor-negative} pair. 
We use a triplet-based network architecture with the pairwise ranking model to learn image
similarity for the triple-based ranking loss function, inspired from~\cite{wang2014learning}. If we have a set of $F = {f_{1},....f_{F}}$ images and $s_{i,j} = s(f_i, f_j)$ that gives the pairwise similarity score between the images $f_i$ and $f_j$. The score $s$ is higher for more similar images and is lower for more dissimilar images. If we have a triplet $t_i = (f_{iA}, f_{iP}, f_{iN})$ where $f_{iA}$, $f_{iP}$ and $f_{iN}$ are the anchor, positive and negative images, respectively. The goal of the training is to learn an embedding function such that:
\begin{equation}
\begin{split}
D(f_{iA}, f_{iP}) < D(f_{iA}, f_{iN}), s(f_{iA}, f_{iP}) > s(f_{iA}, f_{iN})
\label{eq:1}
\end{split}
\end{equation}
where $D(.)$ is the squared Euclidean distance in the embeddings space. A triplet incorporates a relative ranking based on the similarity between the anchor, positive and the negative images. 

The triplet ranking loss function is given as:
\begin{equation}
\label{eq:2}
l(f_{iA}, f_{iP}, f_{iN}) = max\{0, M + D(f_{iA}, f_{iP})-D(f_{iA}, f_{iN})\}
\end{equation}
where $M$ is a parameter called \textit{margin} that regulates the gap between the pairwise distance: $(f_{iA}, f_{iP})$ and $(f_{iA}, f_{iN})$. The model learns to minimize the distance between more similar images and maximize the distance between the dissimilar ones. Our model is based on the work proposed in~\cite{wang2014learning} with the difference that the input image triplets we use are the concatenated embeddings of the masked foregrounds and backgrounds. 


\section{Experiments}

\textbf{Training data.}
\label{sec:training_data}
We use ScanNet dataset~\cite{dai2017scannet} for our experiments which consists of $1500$ indoor RGBD scans annotated with 3D camera poses, surface reconstructions, and mesh segmentation related to several object categories. These annotations allowed us to evaluate the accuracy of Mask-RCNN on the ScanNet images to be used in the proposed pipeline. To generate our training data, we ran Mask-RCNN over a subset of $863$ scenes randomly selected from the whole ScanNet dataset. In total, the Mask-RCNN provided $646,156$ object detections with masks belonging to $29$ object classes (see Table~\ref{tab:scannet_class_freq}). Since not all the objects in the dataset are annotated, we computed the bounding box overlap ratio between the ground truth (GT) bounding boxes and the detections provided by Mask-RCNN to select only the \textit{valid} detections. If the overlap ratio was higher than $60\%$ and the label of the detected object matches with the GT label, it was considered a \textit{valid} detection. 
\begin{table}
\centering
\caption{Number of views after mapping with GT for \textit{valid} detections, selected based on object's label and the bounding box overlap ratio and the number of unique instances for each object category.}
 \begin{tabular}{|c|c|c|c|c|c|} 
 \hline\hline
 \multicolumn{6}{||c||}{No. of views and unique instances per object class} \\
 \hline\hline
 Class & No. of views & No. of instances & Class & No. of views & No. of instances\\ [0.5ex] 
 \hline\hline

bicycle & 110 & 6 & toilet & 1755 & 103\\
\hline
bench & 27 & 4 & tv & 562 & 46\\
\hline
backpack & 1563 & 117 & laptop & 600 & 41\\
\hline
handbag & 486 & 32 & mouse & 59 & 6\\
\hline
suitcase & 377 & 30 & keyboard & 1879 & 67\\
\hline
sports ball & 379 & 21 & microwave & 667 & 61\\
\hline
bottle & 903 & 27 & oven & 72 & 6\\
\hline
cup & 278 & 25 & toaster & 11 & 4\\
\hline
chair & 38203 & 508 & sink & 2694 & 157\\
\hline
couch & 1371 & 75 & refrigerator & 60 & 11\\
\hline
potted plant & 1294 & 55 & book & 3124 & 65\\
\hline
bed & 83 & 17 & clock & 25 & 6\\
\hline
bowl & 121 & 8 & person & 260 & 8\\
\hline
dining table & 1853 & 185 & teddy bear & 47 & 8\\
\hline
vase & 13 & 2 & - & - & -\\
[1ex]
\hline
\end{tabular}
\label{tab:scannet_class_freq}
\end{table}

After mapping each detection obtained from the Mask-RCNN with the corresponding 2D ground truth (GT), we found $9.11\%$ of the total, i.e. around $58876$ detections to be considered fit for the experiments. The regions indicated by the bounding boxes were extended by an additional $10$ pixels-wide border in order to allow loosely-fitted bounding boxes around the objects and thus, allowing a more significant background around each object's mask within the bounding boxes. These regions were then extracted out of the full images, resized to $224\times224$ and categorically stored based on the object's class and it's observed instances. Finally, for each object image, the foreground masks and the background masks were extracted and stored as separate images. The data is split into a 3-fold cross-validation manner with $39250$ images for training and $19626$ images for test over 1701 instances of objects. 

We performed our experiments in three different setups. In all the experimental setups, we used pre-trained ResNet50~\cite{he2016deep} on the ImageNet~\cite{deng2009imagenet} dataset as the backbone model to extract features from the images of the objects. \textbf{no-train:} In this setup, the features extracted from full images were matched against each other by using an $L2$ distance-based metric, without any training. \textbf{full:} In another setup, our model is trained on the embeddings obtained using the full images without extracting separate foreground and background masks. \textbf{concat:} The third type of experimental setup is the approach proposed in this paper where the model is trained on the embeddings obtained by concatenating the features from masked foregrounds and the backgrounds. In \textit{concat} setup, the model learns to minimize the difference between the anchor $f_{iA}$ and the positive $f_{iP}$ images while also learning to maximize the difference between the anchor $f_{iA}$ and the negative $f_{iN}$ images by employing the triplet-loss based training.  
  
\textbf{Evaluation Metrics.}
Most re-ID algorithms use Cumulative Matching Characteristic (CMC) curve as a standard metric to measure their performance which compares the identification rate vs rank. 
The proportions of good matches of the probe image with the set of images in rank-1 would indicate a good or bad performance of the algorithm. A CMC curve is computed for all these individual ranks. In our evaluation procedure, however, we compare with the deepSort~\cite{wojke2017simple} tracking algorithm which is used here as a rank-1 re-ID method, which is why we cannot compare with a CMC curve. Also, it will not be fair to compare recall and precision values between the deepSort and our method. Thus, we compute the rank-1 accuracy by measuring the percentage of correctly identified objects.  

\begin{figure*}[ht]
    \centering
    \includegraphics[width=0.8\textwidth]{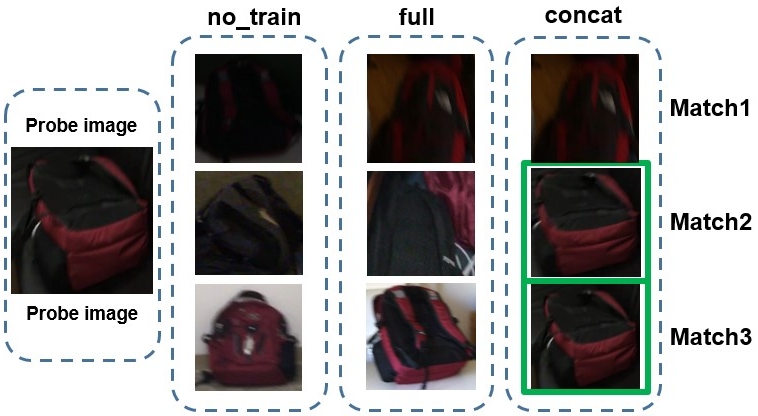}
    \caption[width=0.5\textwidth]{The visualizations show some examples of the matches found in \textit{no-train}, \textit{full} and \textit{concat} setups. The right matches with the probe image are highlighted in green color.}
    \label{fig:vis1}
\end{figure*}

\begin{figure*}[ht]
    \centering
    \includegraphics[width=0.8\textwidth]{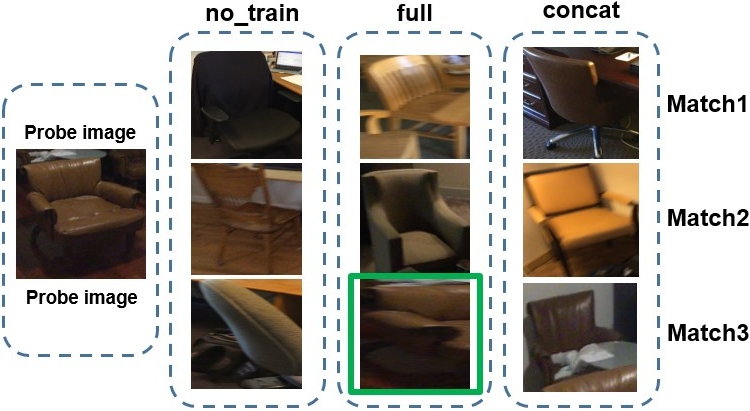}
    \caption[width=0.5\textwidth]{Examples of the matches found in \textit{no-train}, \textit{full} and \textit{concat} setups. The right matches with the probe image are highlighted in green color.}
    \label{fig:vis2}
\end{figure*}

\textbf{Analysis}
Evaluated using the aforementioned experimental setup, the proposed method achieves the best performance on the ScanNet dataset in regards to both the rank-1 accuracy as shown in Table~\ref{table:2}.
Figure~\ref{fig:vis1} shows that the proposed method, \textit{concat} was able to find the best match with the probe image. In the bottom row, \textit{no-train} and \textit{full} tried to match with an image which either had an object of the same color or the shape. However, the proposed method, \textit{concat} could not always correctly identify the images and was performing occasionally poor as can be seen in Figure~\ref{fig:vis2}. Overall, the results from Table~\ref{table:2} show that the \textit{concat} method was able to improve the rank-1 accuracy by $22.19\%$ and $17.1\%$ against \textit{no-train} and \textit{full}, respectively. 
\begin{table}[ht]
    \centering
    \caption{Scores on our ScanNet validation data split with Rank-1,-5,-20 and -50 accuracy values. The best performing type of setups is highlighted in bold.}
    \begin{tabular}{| c | c | c | c | c |} 
     \hline\hline
     \textbf{type} & \textbf{Rank-1}(\%) & \textbf{Rank-5}(\%) & \textbf{Rank-20}(\%) & \textbf{Rank-50}(\%)\\ 
     \hline\hline
    no-train & 55.66 & 66.67 & 77.46 & 89.67 \\
    \hline
    full & 60.75 & 69.61 & 80.90 & 95.21 \\
    \hline
    concat & \textbf{77.85} & \textbf{91.55} & \textbf{98.36} & \textbf{99.80}\\
    [1ex]
    \hline
    \end{tabular}
    \label{table:2}
\end{table}

\begin{figure*}
    \centering
    \includegraphics[width=0.8\textwidth]{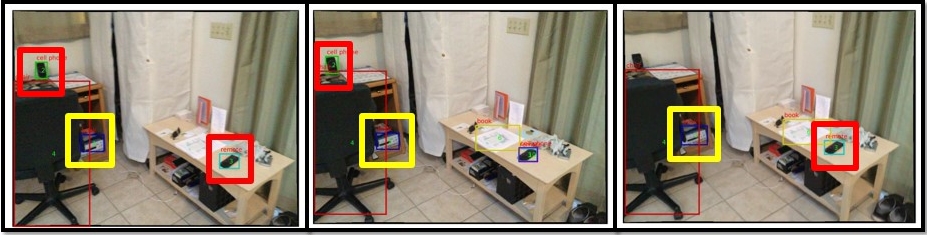}
    \caption[width=0.5\textwidth]{An example object being matched by the deepSort algorithm inside the yellow bounding box and the lost object in the red bounding box.}
    \label{fig:deepsort_vis}
\end{figure*}

\noindent \textbf{Comparison with deepSort}
deepSort~\cite{wojke2017simple} is an open-source implementation of the original SORT~\cite{bewley2016simple} algorithm which employs deep appearance descriptors to improve the performance in multiple object tracking. deepSort learns discriminative feature embeddings offline in order to obtain a deep association metric for a person re-identification dataset in the original work. For our experiments, we provided two random sets of image pairs obtained from the ScanNet scenes to the algorithm to identify multiple objects ensuring that an image pair is not consisting of images from two different scenes. We computed the performance by measuring the percentage of matched object instances across all the image pairs. 
Figure~\ref{fig:deepsort_vis} shows the possible problems that standard object matching or tracking algorithms might face in re-identifying objects. The figure shows that the deepSort was able to match an object (in yellow bounding box) in multiple frames but lost an object (in red bounding box) when the camera revisits a similar view later. deepSort  achieved a rank-1 accuracy of $49.60\%$ against the rank-1 accuracy of $77.85\%$ obtained with our method.



\section{Conclusion}
The contribution of this paper was to explore the intuition that the information obtained from the background surrounding the detected target objects in a rigid scene could be highly useful in discriminating two near-identical objects or two instances of the same object class. The discriminative features learned from the explicit concatenated foreground and background can be utilized to re-identify objects at the instance-level throughout the dataset. Our experiments have shown that the proposed method performs well even in the case of highly cluttered rigid environments like the indoor scenes obtained from ScanNet dataset. In future, we plan to explore if the temporal information obtained from multiple views in a video dataset can be integrated with our object instance re-identification system for a robust multiple object tracking algorithm in case of rigid and static scenes.       
%
%
%
%




\bibliographystyle{splncs04}
\bibliography{refs_arXiv}

\end{document}